\documentclass[runningheads]{llncs}

\usepackage{graphicx}
\usepackage{amssymb}
\usepackage{amsmath}
\usepackage{wrapfig}
\usepackage{multicol}
\usepackage{xcolor}
\usepackage{subcaption}
\usepackage{float}
\usepackage{array}

\definecolor{newGray} {RGB}{50,50,50}
\definecolor{newBlue} {RGB}{128,128,255}
\definecolor{newRed} {RGB}{255,128,128}

\begin{document}

\title{Latent Time-Adaptive Drift-Diffusion Model}

\author{Gabriele Cimolino\inst{1}\orcidID{0000-0003-4832-0874} \and \\ 
Fran\c{c}ois Rivest\inst{1,2}\orcidID{0000-0003-2038-5174}}

\authorrunning{G. Cimolino \& F. Rivest.}

\institute{Queen's University, Kingston ON, Canada\\
\email{gabriele.cimolino@queensu.ca} \and
Royal Military College of Canada, Kingston ON, Canada\\
\email{francois.rivest@rmc.ca}}

\maketitle

\begin{abstract}
Animals can quickly learn the timing of events with fixed intervals and their rate of acquisition does not depend on the length of the interval. In contrast, recurrent neural networks that use gradient based learning have difficulty predicting the timing of events that depend on stimulus that occurred long ago. We present the latent time-adaptive drift-diffusion model (LTDDM), an extension to the time-adaptive drift-diffusion model (TDDM), a model for animal learning of timing that exhibits behavioural properties consistent with experimental data from animals. The performance of LTDDM is compared to that of a state of the art long short-term memory (LSTM) recurrent neural network across three timing tasks. Differences in the relative performance of these two models is discussed and it is shown how LTDDM can learn these events time series orders of magnitude faster than recurrent neural networks.
\end{abstract}

\section{Introduction}
Animals are able to quickly learn the timing of events with fixed time intervals~\cite{balsam_timing_2002} and the rate at which these timings are learned is independent of the time interval length to learn~\cite{gibbon_scalar_1977}. This property of animal learning of timing is called \textit{timescale invariance} and it means that animals can learn the timing of events in the same number of attempts, as long as the ratio between intertrial intervals and interstimulus intervals is the same~\cite{gallistel_time_2000}. This ability is necessary for performing real world tasks, where the timing of actions is important, but it is one that many deep learning models do not have. As the length of the interval to be learned grows, state of the art deep learning models can require an exponential number of attempts to learn what animals can in just a few~\cite{BlumRivest1992}.

Recurrent neural networks, such as spiking neural networks (SNN)~\cite{trappenberg_fundamentals_2010} and long short-term memory (LSTM)~\cite{hochreiter_long_1997}, are capable of modelling the timing of events with fixed intervals but have model specific difficulties in learning them. SNNs often have activation functions that are not differentiable, making them incapable of learning using gradient descent~\cite{tavanaei_deep_2019}. LSTMs learn time series data by minimizing their instantaneous prediction error or loss at each discrete time step~\cite{hochreiter_long_1997}. This enables them to learn temporal associations between conditioned and unconditioned stimuli; however their memory mechanisms may not hold critical stimuli that occurred long ago~\cite{gers_learning_2003}. Their ability to learn timing depends on the length of the interval between events, which can prevent them from learning long-term associations.

Prior work has posited a model for animal learning of interval timing, called the time-adaptive drift-diffusion model (TDDM)~\cite{luzardo_driftdiffusion_2017,rivest_adaptive_2011,Simenetal2011}, that exhibits properties observed in animal data. TDDM works by learning the evidence rate that stimulus are providing about an upcoming events and by accumulating those evidences to predict the timing of an event. A major limitation of this fast learning model, is that it is made of a single output layer of temporal accumulators, limiting its capacity to problems with linear solutions in the stimulus duration space. In this paper, we present an extension to TDDM by creating a hidden layer of latent timed variables and developing a gradient over the squared timing error (STE) \cite{rivest_new_2020} of the output units with respect to the latent variable timing and internal accumulation process. This allows the network to learn latent or hidden timed features that can be used to improve the network timing output.

In the next section, we first discuss recent approaches to learning timing with deep neural networks. We then summarize the TDDM approach in section \ref{sec:TDDM}. Then in section~\ref{sec:LTDDM} we describe how to extend it to support a single hidden layer, acting as latent timed variables. In section~\ref{sec:exp} we compare our new latent time-adaptive drift-diffusion model (LTDDM) to state of the art LSTM on 3 timing prediction datasets: heartbeats, music, and stock predictions, followed by a discussion. 

\vspace{-0.25cm}
\section{Related Work}\label{sec:RelW}
In this section we present recent approaches to enabling deep neural models to learn the timing of events. These other approaches have sought to leverage dynamic time warping (DTW) and reinforcement learning to improve the predictive power of deep and recurrent neural networks. We will address these to explain their similarities and differences with the present work.

DTWNet is a neural architecture that uses a temporal loss function to learn feature extractors via backpropagation~\cite{cai_dtwnet_2019}. The temporal feature extraction approach was inspired by the convolution kernels of convolutional neural networks, which learn both local and global spatially associated features. Using a differentiable form of DTW, DTWNet learns fixed-length DTW kernels that extract local and global temporally associated features in time series. Although both DTWNet and LTDDM learn temporal features latent in time series, LTDDM's features are discrete events evidenced by stimulus observed since their last activation reset. This means that, while LTDDM is only capable of predicting the timing of events and not their amplitude, the timing features that it learns can have arbitrarily long temporal associations.

DILATE is a temporal loss function used to train deep learning models for multi-step prediction in sequence to sequence translation~\cite{le_guen_shape_2019}. It features two DTW inspired error terms that regulate the shape and timing of the network's predictions. Using DILATE, deep and recurrent models have been trained to reproduce synthetic step function and electrocardiogram time series more accurately than networks trained using mean squared error. Because the time series of interest to the present work are discrete, STE is better suited to our purpose than DILATE.

Deverett et al. investigated recurrent neural networks' abilities to reproduce observed timing intervals in an experimental setting~\cite{deverett_interval_2019}. They found that LSTM was capable of nearly perfectly reproducing sample intervals when trained using an actor-critic architecture. Networks were tasked with reproducing the interval between two events immediately after the occurrence of the second. This means that, unlike how TDDMs are supervised to produce only a single fixed interval through repeated exposures, the LSTMs used in this experiment learned a general rule for interval mimicry, by encoding the duration of their most recently observed interval in their internal states. Determining whether recurrent neural networks can learn timing rules that encompass temporal associations between multiple intervals was outside the scope of this work, but this experimental setup represents a promising approach to training predictive models with reinforcement learning to develop general rules for complex interval timing.

Finally, Music Transformer takes a natural language processing approach to learning the timing of notes in music~\cite{huang_music_2018}. Just as the Transformer architecture leverages spatial correlations in the ordering of words in an utterance, Music Transformer extracts temporal correlations among the notes in a piece of music. This enables it to discover latent features corresponding to musical structures (e.g., scales, arpeggios, sections) at different time scales. From this perspective, the theoretical foundations of Music Transformer and the present work are similar (i.e., the structure of time series data can be learned at multiple levels of abstraction). However, Music Transformer, as with other Transformers, learns by reducing its instantaneous error at each time step since its loss function does not express a timing error. When run in an autoregressive manner, neural networks using Music Transformer have been shown to produce coherent piano music up to one minute in length~\cite{hawthorne_enabling_2019}.

\section{Background}\label{sec:TDDM}
\begin{figure}
    \begin{subfigure}{\linewidth}
        \hspace{0.3\linewidth}\fcolorbox{black}{newGray}{\rule{0pt}{4pt}\rule{4pt}{0pt}}\quad Ground Truth
        \hspace{0.05\linewidth}\fcolorbox{black}{newRed}{\rule{0pt}{4pt}\rule{4pt}{0pt}}\quad Output
    \end{subfigure}\par\medskip
    \begin{subfigure}{\linewidth}
        \includegraphics[width=\textwidth]{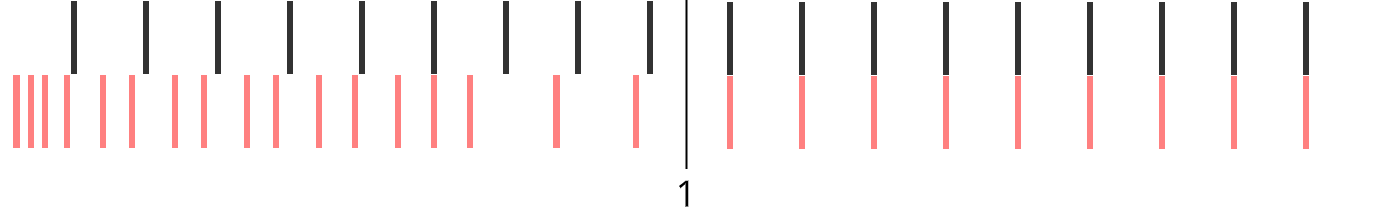}
    \end{subfigure}\par\medskip
    \caption{A TDDM learning a target sequence with a single fixed interval over two epochs.} \label{fig:FI}
\end{figure}

The time-adaptive drift-diffusion model (TDDM) is a model of animal learning of timing that learns drift rates that attribute evidence to conditioned stimuli. With constant stimulus, a TDDM is able to learn a single fixed interval between events. More complicated sequences are learned by approximating the mean length of multiple intervals. The rate at which intervals are learned is a function of the number of observations and does not depend on the length of the intervals. The model's rate of acquisition is timescale invariant, enabling the TDDM to quickly learn sequences of target events with a single fixed interval and approximate complex sequences, with composite period given appropriate stimuli, such as the occurrences of notes in music. An example of TDDM's quick learning ability is shown in Fig.~\ref{fig:FI}.

\subsubsection{Sequences} Let $x_t\in \{0,1\}^{N}$ be the $N$-dimensional binary observation at time step $t$, where $x_t$ first component is a bias ($x_{t,1} = 1$), and let $y_t\in \{0,1\}$ be the desired target output at time step $t$.

\subsubsection{Accumulators} In TDDM, the new observed stimuli $x_t$ are added to its accumulators $a_t \in \mathbb{N}^{N}$ such that $a_t = a_{t-1} + x_t$. In this way, a TDDM integrates stimuli indicative of the next target event.

\subsubsection{Drift Rates} A TDDM attributes evidence to its observed stimuli according to drift rates $w_t \in \mathbb{R}_{>0}^{N}$. Drift rates are TDDM's learnable parameters similar to artificial neural networks weights. Negative stimulus is conceptually undefined, so drift rates are constrained within the positive reals to guarantee convexity in the error surface.

\subsubsection{Activation} At each time step, a TDDM's accumulated evidence is summed to give its total activation $\phi_t = w_t \cdot a_t$. A TDDM's output $\hat{y_t} \in \{0,1\}$ is set according to 
\[\hat{y_t} = \begin{cases}
    1,& \text{if } \phi_t \geq \tau\\
    0,& \text{otherwise}
\end{cases}\]
given an arbitrary activation threshold $\tau$. For simplicity, we assume $\tau = 1$. In this way, $\hat{y_t}$ is defined by the hyperplane $w_t \cdot a_t = \tau$, beyond which TDDM`s output is activated. Upon activation, a TDDM's integrator $a_t$ is reset to $a_t = \{0\}^{N}$ and the accumulation of stimulus for the next target event begins again.

\subsubsection{Timing Error} TDDM's error gradient is different than similar neural models. Neural networks typically reduce their instantaneous error or loss at each time step in order to learn sequences in time. Their predictions have variable extension in a prediction space that are brought closer to desired values by descending the resulting error gradient. This can be thought of as a kind of \textit{amplitude error}, because they reduce the amplitude difference between their prediction and the target value at each time step. TDDM on the other hand reduces differences in the time at which it predicts events and the time at which they actually occur. They can be said to have \textit{timing error}, which is reduced on-line as target events occur.

Should $y_t \neq \hat{y_t}$ then a TDDM has incurred some error in its prediction. Rather than minimize the model's instantaneous error, as is necessary with other neural networks, TDDM performs gradient descent on the gradient of its squared timing error (STE) with respect to its drift rates. The STE for an event in a given sequence is the squared difference in the time at which an event occurred in that sequence and the closest event that occurred in the other sequence~\cite{rivest_new_2020}. When considering the entirety of both sequences $Y = (y_t)$ and $\hat{Y}= (\hat{y}_t)$, the STE is
\begin{equation*}
    \begin{split}
        STE(Y, \hat{Y}) = \frac{1}{2} \sum_{t = 0}^{T} \min_{\hat{t}} (t - \hat{t} \vert y_{t} = 1, \hat{y}_{\hat{t}} = 1)^2 \\ + \frac{1}{2} \sum_{\hat{t} = 0}^{T} \min_{t} (\hat{t} - t \vert y_{t} = 1, \hat{y}_{\hat{t}} = 1)^2.
    \end{split}
\end{equation*}

While in prior work~\cite{rivest_new_2020} the STE of both the target sequence and the model's prediction were minimized, in this paper we correct the network only when a target event occurs, as was done in~\cite{luzardo_2013} and~\cite{rivest_adaptive_2011}. This can be seen as taking only one term of the STE to get the loss:

\[STE_{t}(Y, \hat{Y}) = \frac{1}{2} \min_{\hat{t}} (t - \hat{t} \vert y_t = 1, \hat{y}_{\hat{t}} = 1)^2.\]

\subsubsection{Correction} The timing error gradient of a TDDM is the gradient of its STE with respect to $\epsilon \vert \hat{y_{\epsilon}} = 1$, the time at which it becomes activated. This is minimized by descending the gradient of its STE with respect to its drift rates. To do this we approximate its accumulated stimuli at time $t$ by scaling its observed stimuli according to $\lambda_{t} = \frac{a_{t,1} - \nabla_{\epsilon} STE_{t}(Y, \hat{Y})}{a_{t,1}}$, where $a_{t,1}$ is the accumulator that corresponds with the bias event indicating time elapsed since its last activation. A TDDM's timing is corrected to minimize its STE by translating its drift rates to be closer to the activation hyperplane $w_{t} \cdot \lambda_{t} a_{t} = \tau$. This shifts the timing of the unit's activation to be $\nabla_{\epsilon} STE_{t}(Y, \hat{Y})$ time steps later. Therefore, the gradient of its STE with respect to its drift rates is defined as

\[ \nabla_{w_{t}} STE_{t}(Y,\hat{Y}) = (w_t \cdot  \lambda_{t} a_{t} - \tau) \frac{\lambda_{t} a_{t}}{\|\lambda_{t} a_{t}\|_2^2}.\]

Rather than applying a correction at the occurrence of each event in $Y$ and $\hat{Y}$, we suppress the TDDM's activation until the occurrence of an event in $Y$ and apply the correction then. By forcing TDDM to become activated at the occurrence of an event $Y$, it incurs no timing error and $\lambda_t$ is 1. This is simpler, yielding the gradient

\[ \nabla_{w_{t}} STE_{t}(Y, \hat{Y}) = (w_{t} \cdot a_{t} - \tau) \frac{a_{t}}{\|a_{t}\|_2^2}.\]

\section{Latent Time-Adaptive Drift-Diffusion Model}\label{sec:LTDDM}
In this paper, we propose the latent time-adaptive drift-diffusion model (LTDDM), an extension to the TDDM that enables it to learn target sequences with multiple fixed intervals given constant stimulus. In particular, TDDM is unable to learn complex target sequences such as the occurrences of notes in music. This is because the prediction of target events may depend on stimulus that occurred further back in time than the last target event. Given that humans are capable of learning the timing of such target sequences quickly, we have extended TDDM to include a hidden TDDM layer, acting as timed latent stimuli, enabling it to learn functions of composite period in the presence of constant stimulus.

\begin{figure}
    \begin{subfigure}{\linewidth}
        \hspace{0.3\linewidth}\fcolorbox{black}{newGray}{\rule{0pt}{4pt}\rule{4pt}{0pt}}\quad Ground Truth
        \hspace{0.05\linewidth}\fcolorbox{black}{newRed}{\rule{0pt}{4pt}\rule{4pt}{0pt}}\quad Output
    \end{subfigure}\par\medskip
    \begin{subfigure}{\linewidth}
        \includegraphics[width=\textwidth]{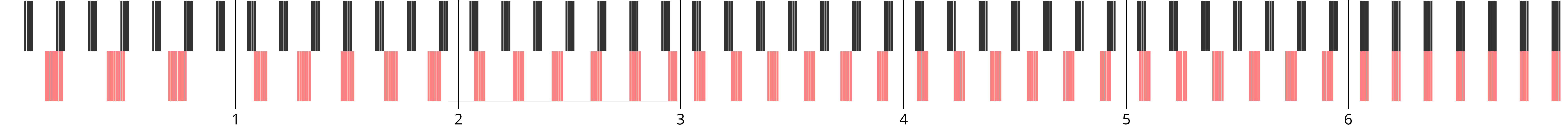}
    \end{subfigure}\par\medskip
    \caption{A LTDDM learning a target sequence with a fixed interval and a single duration over seven epochs. The unit initially predicts an interval (OFF) and duration (ON) that are too long. By reducing its timing error, the LTDDM learns to perfectly predict the fixed interval ON and OFF times.} \label{fig2}
\end{figure}

\vspace{-1cm}
\subsubsection{Activation and Deactivation Timing} While basic TDDM produces behaviours by simply timing an instantaneous upcoming event, in order to generate hidden or latent stimuli, we need units to remain in the ON or OFF state for a certain amount of time. Fig.~ \ref{fig2} is a visualization of an LTDDM learning the duration of ON and OFF events in a target sequence. In \cite{rivest_new_2020}, they model ON and OFF state of music notes simply by considering turning ON events, and turning OFF events, independently. This approach does not work for hidden units, as it is important to considered their ON period as a whole. Thus, in this paper, we extend TDDM units to be bi-stable ON/OFF units instead of units producing instantaneous events. Each unit will thus be made of two accumulators, one accumulating evidence during the OFF state, triggering the ON state, and one accumulating evidence during the ON state, to trigger the OFF state. 

This modification provides several benefits to learning complex sequences of events. LTDDMs are capable of learning a richer set of functions, wherein target events occur for longer than a single time step. They are also able to have their activation and deactivation timings modified to match behaviour desired by TDDM units in successive layers. In particular, it provides greater control over hidden unit's length of ON state. This enables units upstream to modify the duration of hidden units to be more predictive of events in the target sequence.

\begin{wrapfigure}{r}{0.45\textwidth}
    \centering
    \vspace{-0.75cm}
    \includegraphics[width=0.44\textwidth]{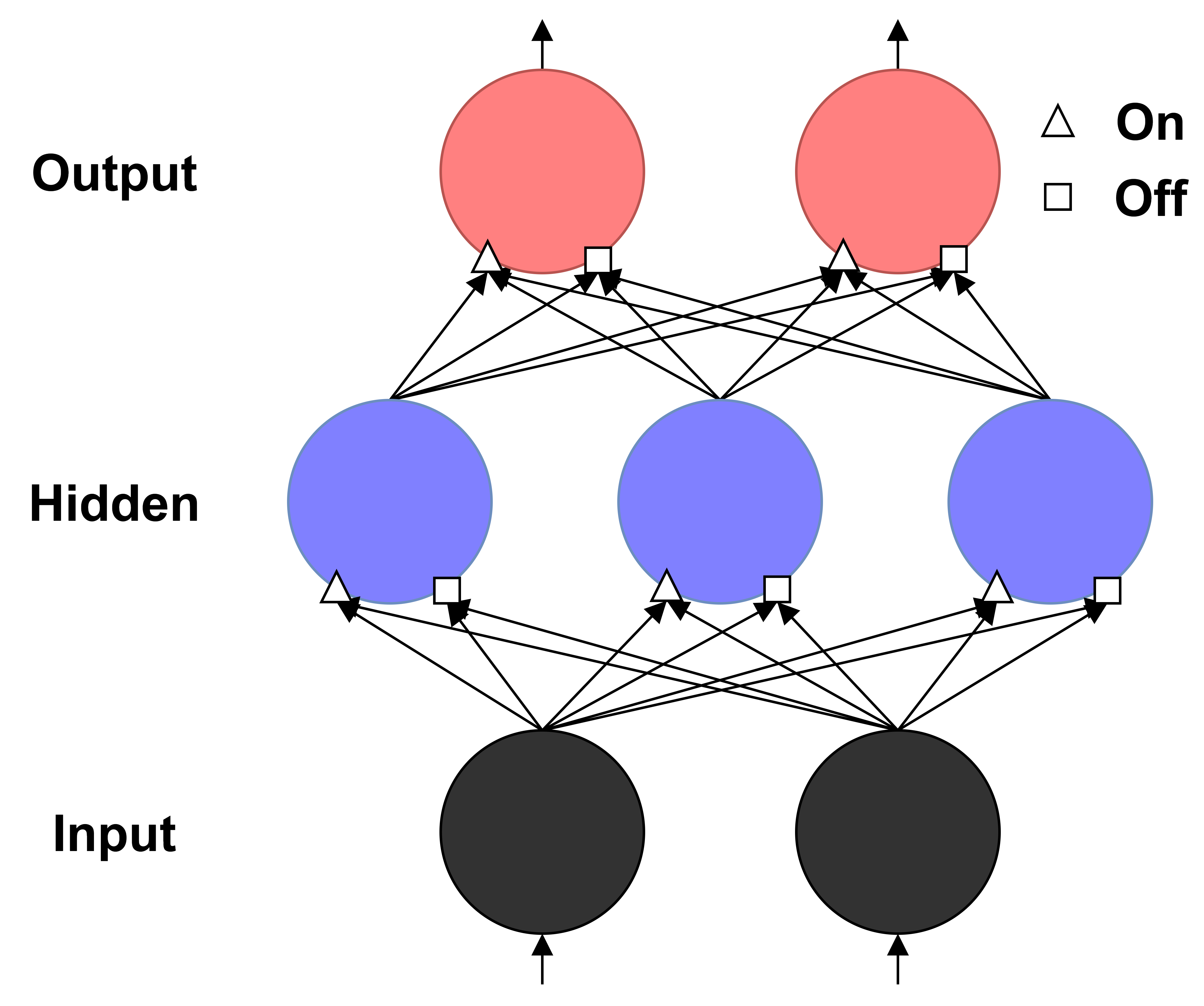}
    \caption{A LTDDM network with one output layer and one hidden layer.} \label{fig3}
    \vspace{-0.75cm}
\end{wrapfigure}

\subsubsection{Network Structure and Dynamics} A LTDDM is a fully connected, feed forward network composed of layers of TDDM units that learn the timing and duration of observed and hidden events. When a LTDDM unit reaches its activation threshold, it switches state from on to off or off to on. Upon entering a state, its accumulators are reset to begin collecting evidence for the next target event. Corrections to each unit's timing and duration can only be made with respect to observations made during its current and previous states. A benefit of retaining observations made in a previous state is the ability to modify both the timing and duration of predicted events. Figure \ref{fig3} is a diagram of a small example LTDDM network.

\subsubsection{Timing Error Gradients} When an event occurs in the target sequence, the timing and duration of the LTDDM's closest prediction are modified to match the target event. This is done by correcting the network's output layer with respect to its evidence rates, as is done with TDDMs, and propagating the sum of the output units' timing error gradients, with respect to their accumulated stimuli, downstream to the previous layer. Conceptually, the gradient of a LTDDM's STE with respect to $a_t$ is a dual of the gradient of its STE with respect to $w_t$.

\vspace{-1cm}
\begin{multicols}{2}
    \begin{equation*}
        \resizebox{0.9\hsize}{!}{
        $\nabla_{w_{t}} STE_{t}(Y, \hat{Y}) = (w_{t} \cdot a_{t} - \tau) \frac{a_{t}}{a_{t} \cdot a_{t}}$
        }
    \end{equation*}\break
    \begin{equation*}
        \resizebox{0.9\hsize}{!}{
        $\nabla_{a_{t}} STE_{t}(Y, \hat{Y}) = (w_{t} \cdot a_{t}- \tau) \frac{w_{t}}{w_{t} \cdot w_{t}}$
        }
    \end{equation*}
\end{multicols}

Given arbitrary $w_t$ and $a_t$, there exist hyperplanes of $w_t$ and $a_t$ where $\phi_t = \tau$ and $STE_t = 0$. Where $\nabla_{w_{t}} STE_{t}(Y, \hat{Y})$ says that a unit would have predicted correctly had it attributed a different amount of evidence to its observations, $\nabla_{a_{t}} STE_{t}(Y, \hat{Y})$ says that a unit would have predicted correctly had it seen a different amount of observations. These gradients are summed and passed to the previous layer so that its units can correct their behaviour to provide the desired amount of stimulus. 

\subsubsection{Hidden Unit Correction} 
\begin{figure}
    \includegraphics[width=\textwidth]{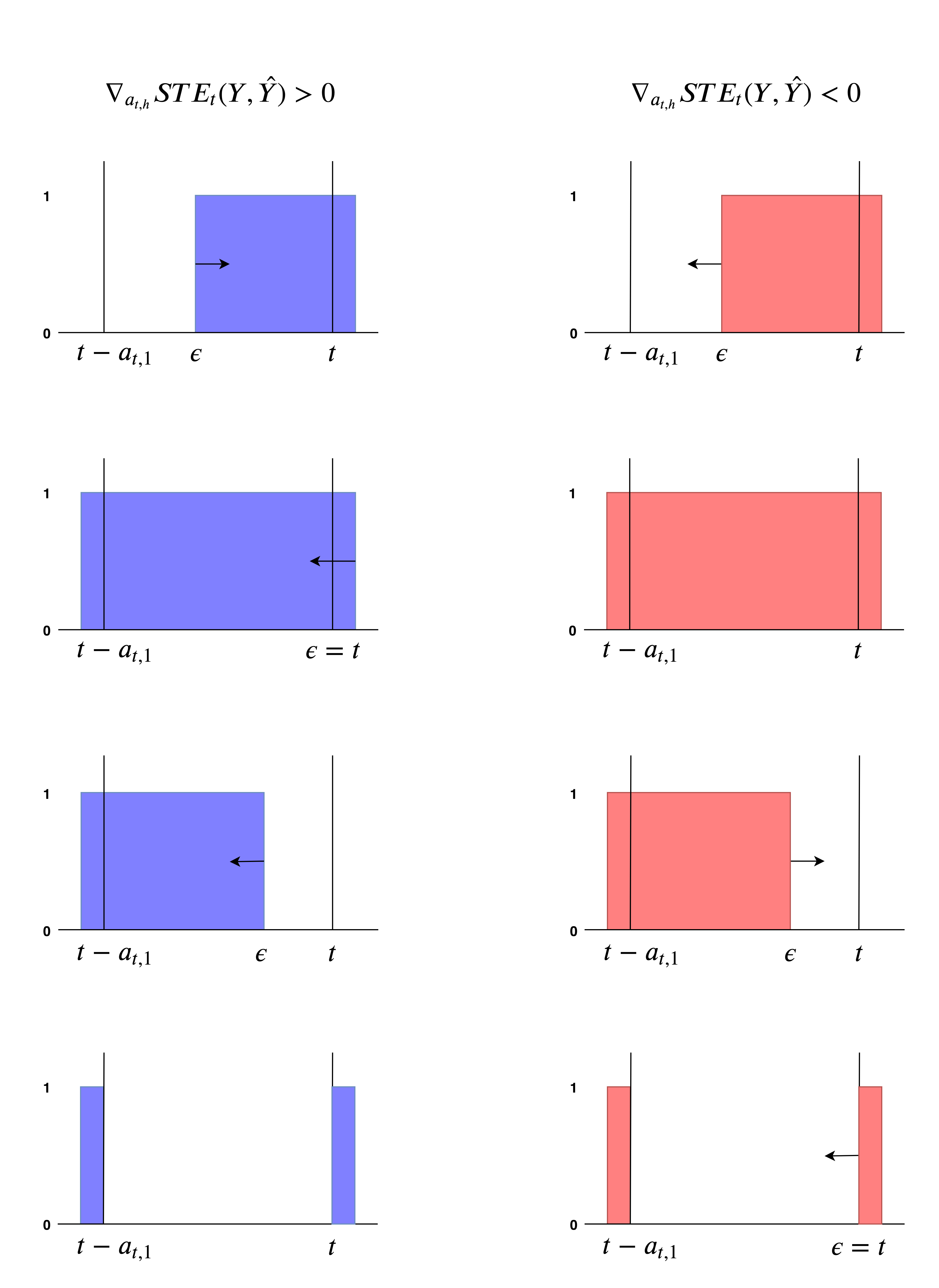}
    \caption{The LTDDM correction rules applied to a hidden unit depending on its recent behaviour and the output unit's desired change. The block (blue or red) represents the hidden unit's state. $t-a_{t,1}$ and $t$ represents the start and stop time of the state of the clamped output unit. Blue indicates that the output unit's timing error gradient is positive (shorter latent stimulus is better); red indicates that it is negative (longer latent stimulus is better). At time step $t$, the output unit is corrected and sends a correction request to the hidden units. $\epsilon$ is the time of the hidden unit's state boundary that the hidden unit will attempt to change in the direction of the arrow by changing its weights to improve the output unit's behaviour.} \label{fig1}
\end{figure}

When correcting a hidden unit, both the timing and duration of its prediction can be modified. If the timing gradient with respect to a given unit's behaviour is negative, indicating that the successive layer needs more stimuli, the unit can be corrected by either having it turn on sooner or turn off later. Conversely, if it is positive, a desire for fewer stimuli can be accommodated by having the unit turn on later or turn off earlier. Whether the unit's timing or duration is modified depends on its recent behaviour from the perspective of the units upstream. LTDDM uses simple learning rules that consider changes in state of the units downstream within a time window since the upstream unit's last change in state, as seen in Fig~\ref{fig1}. Where $\epsilon$ is the time at which a hidden unit with index $h$ changes state to be either ON or OFF, the gradient of the output unit's STE with respect to $\epsilon$, for both ON and OFF timing, is defined as:

\vspace{0.5cm}
\resizebox{0.9\textwidth}{!}{
$
\nabla_{\epsilon \vert \phi^{h, ON}_{\epsilon} \geq \tau} STE_{t}(Y, \hat{Y}) = 
\begin{cases}
    \nabla_{a_{t, h}} STE_{t}(Y, \hat{Y}),& \text{if } a_{t, h} < a_{t, 1} \text{ and } \hat{Y^h} = 1 \\
    \nabla_{a_{t, h}} STE_{t}(Y, \hat{Y}),& \text{if } a_{t, h} = 0 \text{ and } -\nabla_{a_{t, h}} STE_{t}(Y, \hat{Y}) > 0
\end{cases}
$
}

\vspace{0.5cm}
\resizebox{0.9\textwidth}{!}{
$
\nabla_{\epsilon \vert \phi^{h, OFF}_{\epsilon} \geq \tau} STE_{t}(Y, \hat{Y}) = 
\begin{cases}
    -\nabla_{a_{t, h}} STE_{t}(Y, \hat{Y}),& \text{if } a_{t, h} < a_{t, 1} \text{ and } \hat{Y^h} = 0 \\
    -\nabla_{a_{t, h}} STE_{t}(Y, \hat{Y}),& \text{if } a_{t, h} = a_{t, 1} \text{ and } -\nabla_{a_{t, h}} STE_{t}(Y, \hat{Y}) < 0
\end{cases}
$
}
\vspace{0.5cm}

Unlike the output unit whose state is clamped to its target signal, latent or hidden unit states are running freely, switching state when their accumulator reaches threshold. When the output unit requests a correction to a hidden unit that has changed state within the window of time since the output unit's last change, only the timing of the hidden unit's last change in state is modified by modifying its weights. When the hidden unit has not changed state within the window, only the timing of its next change in state is modified. There is an exception when fewer observations are desired but there were none; or more observations are desired but the hidden unit was on for the entire window, in which case no correction is applied.

\begin{figure}
    \begin{subfigure}{\linewidth}
        \hspace{0.2\linewidth}\fcolorbox{black}{newGray}{\rule{0pt}{4pt}\rule{4pt}{0pt}}\quad Ground Truth
        \hspace{0.05\linewidth}\fcolorbox{black}{newRed}{\rule{0pt}{4pt}\rule{4pt}{0pt}}\quad Output
        \hspace{0.05\linewidth}\fcolorbox{black}{newBlue}{\rule{0pt}{4pt}\rule{4pt}{0pt}}\quad Hidden
    \end{subfigure}\par\medskip
    \begin{subfigure}{\linewidth}
        \includegraphics[width=\textwidth]{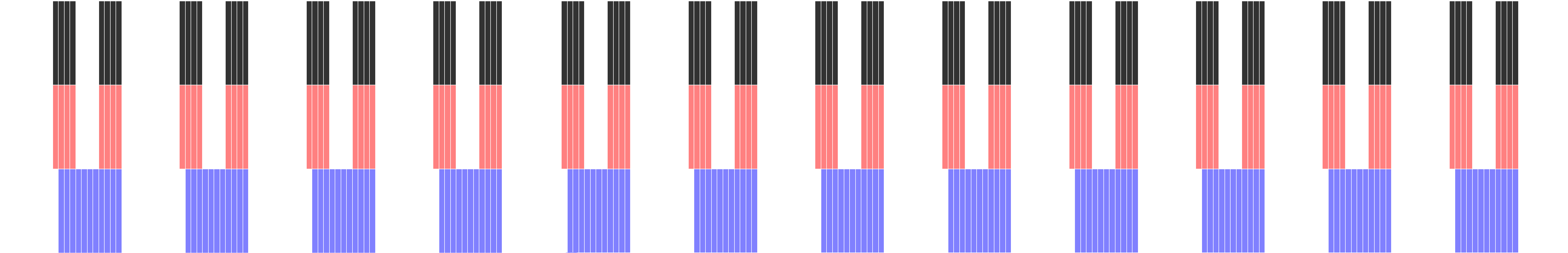}
    \end{subfigure}\par\medskip
    \caption{A LTDDM, with 2 layers of 1 unit each, that has learned a target sequence with two fixed intervals and a single duration, shown in black. The network's hidden unit, shown in blue, has been corrected to stay on for the duration of both target events and the short interval.} \label{fig:LTDDM}
\end{figure}

If there are multiple output units, each hidden unit receives correction signals through each output units' error gradients with respect to its own timing. Hidden units then apply corrections by shifting their timing to provide the amount of stimulus desired by units upstream. Fig.~\ref{fig:LTDDM} shows the activity of an LTDDM network that has learned to predict a target sequence with two ON interval lengths and two OFF interval lengths.

\section{Experiment}\label{sec:exp}
We compared the performance of TDDM and LTDDM against that of LSTM across three sets of time series data with binary features. These data represented a human heartbeat, the change in stock market prices, and the occurrence and duration of notes in music. They were chosen to vary in sparsity and rhythmic structure.

LSTM represents the state of the art in recurrent neural networks. It was chosen to provide a comparison between LTDDM, which reduces its error in time, and recurrent neural networks, which reduce their error in output at each time step. For each data set, we used the Keras implementation of a single layer LSTM network, trained by reducing its binary cross entropy using the Adam optimizer with learning rate 0.001, $\beta_1$ 0.9, and $\beta_2$ 0.999. Its performance was compared to a LTDDM, corrected on-line with learning rate 0.1, with one hidden layer of as many units as there were outputs. At each time step, both networks were presented with the previous time step of the target sequence as input in an autoencoder fashion. LTDDM's output layer received both the network's inputs and the outputs of its hidden layer. The purpose of this experiment was to test the abilities of both models to reproduce the observed sequences, so generalization to other sequences was not investigated. TDDM and LTDDM were trained for 200 epochs and LSTM was trained for 1000 epochs on all tests, where an epoch is one pass through the sequence from beginning to end. The number of training epochs were chosen because early testing indicated that the models' rates of improvement decreased significantly before these epochs. The predictions of LTDDM and LSTM after 200 epochs of training on all three data sets are shown in Fig.~\ref{fig:samples} and in the appendix. We will now describe the three time series used in our experiment.

\vspace{-0.5cm}
\subsubsection{Heartbeat}
The heartbeat data were digital electrocardiogram signals drawn from the MIT-BIH Normal Sinus Rhythm Database (nsrdb)\footnote{MIT-BIH Normal Sinus Rhythm Database is available from www.physionet.org/content/nsrdb/1.0.0/.}. Ten of the eighteen traces\footnote{Traces 16265m, 16272m, 16273m, 16420m, 16483m,
16539m, 16786m, 17052m, 17453m, and 18184m were used.} were analyzed using MATLAB's findpeaks function to identify peaks that crossed a 0.5 mV threshold as heartbeats~\cite{rivest_new_2020}.
\vspace{-0.5cm}
\subsubsection{Music} The musical piece used in this experiment was \textit{Die Naehterin's sass a Naehterin und sie naeht}, taken from the Essen Folksong Collection\footnote{Essen Folksong Collection kern scores are available from www.kern.humdrum.org/cgi-bin/browse?l=/essen.}. The song's kern data was parsed as a list of tone indices and durations. Tones were indexed from 0 to 96 corresponding to 8 octaves of 12 tones, and their durations were measured in 32\textsuperscript{nd} notes. The tones were folded into a single octave of 12 tone sequences with 32\textsuperscript{nd} note time steps. Sequences with wherein no notes occurred end with a single an event on the last time step, so that all sequences have at least one event.
\vspace{-0.5cm}
\subsubsection{Stock}
Daily closing price data for all NASDAQ-100 companies\footnote{Closing price data are available from www.nasdaq.com}, between April 1, 2013 and March 31, 2014, were discretized to indicate large jumps in price day to day~\cite{rivest_new_2020}. Target events represent closing prices at least 0.7 standard deviations greater than the mean price of the last 14 days.

\section{Results}
An analysis of the predictions of the three models across all three data sets highlights their behavioural characteristics, as seen in the table above. LTDDM is verbose; it makes many predictions, often multiple for the same target event, producing much higher recall when target events occur infrequently. In contrast, LSTM is terse and precise. When presented with temporally sparse data, LSTM prefers to err on the side of the sequence mean and remain silent. 

\begin{figure}[b!]
    \setlength{\fboxsep}{0pt}
    \setlength{\fboxrule}{1pt}
    \begin{subfigure}{\linewidth}
        \hspace{0.2\linewidth}\fcolorbox{black}{newGray}{\rule{0pt}{4pt}\rule{4pt}{0pt}}\quad Ground Truth
        \hspace{0.05\linewidth}\fcolorbox{black}{newRed}{\rule{0pt}{4pt}\rule{4pt}{0pt}}\quad LTDDM
        \hspace{0.05\linewidth}\fcolorbox{black}{newBlue}{\rule{0pt}{4pt}\rule{4pt}{0pt}}\quad LSTM
    \end{subfigure}\par\medskip
    \begin{subfigure}{0.32\linewidth}
        \fbox{\includegraphics[width=\textwidth]{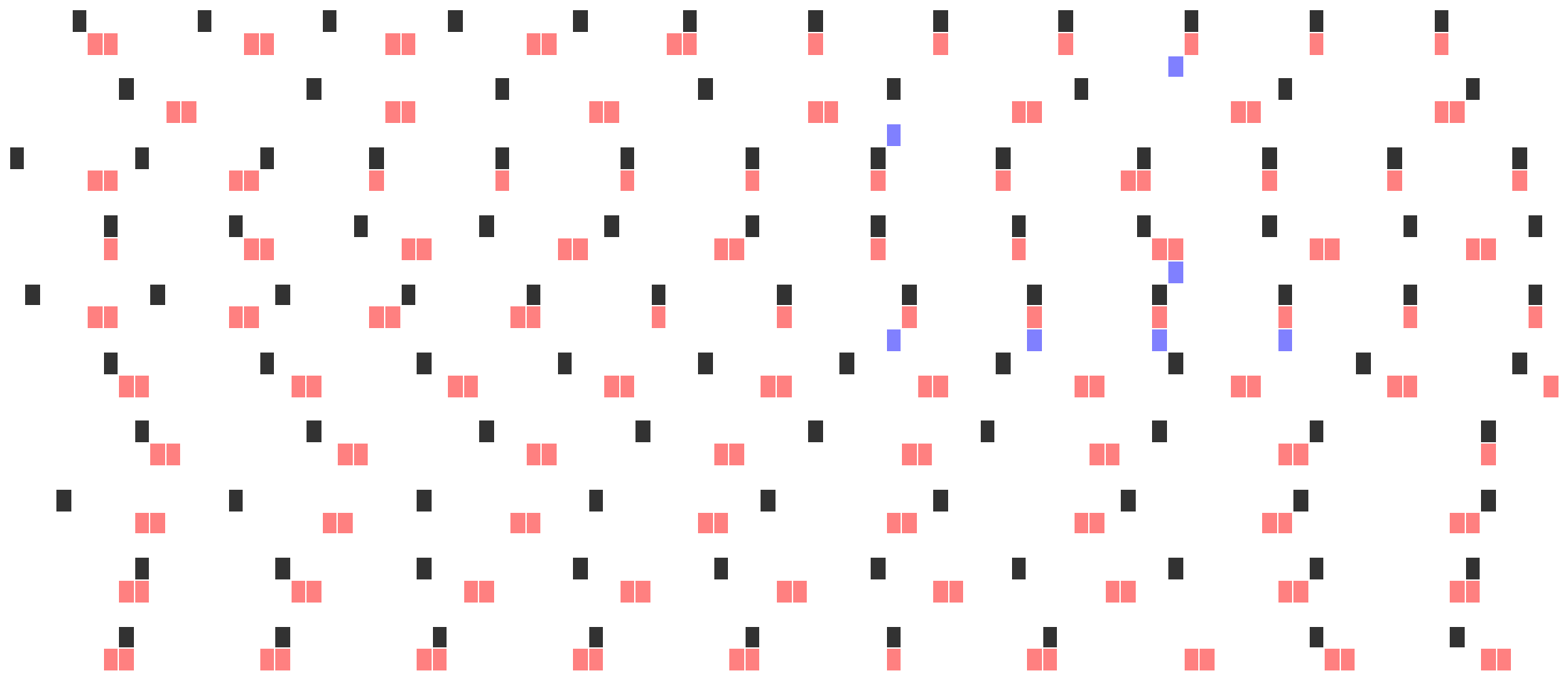}}
        \caption*{Heartbeat}
    \end{subfigure}\hspace{0.01\linewidth}
    \begin{subfigure}{0.32\linewidth}
        \fbox{\includegraphics[width=\textwidth]{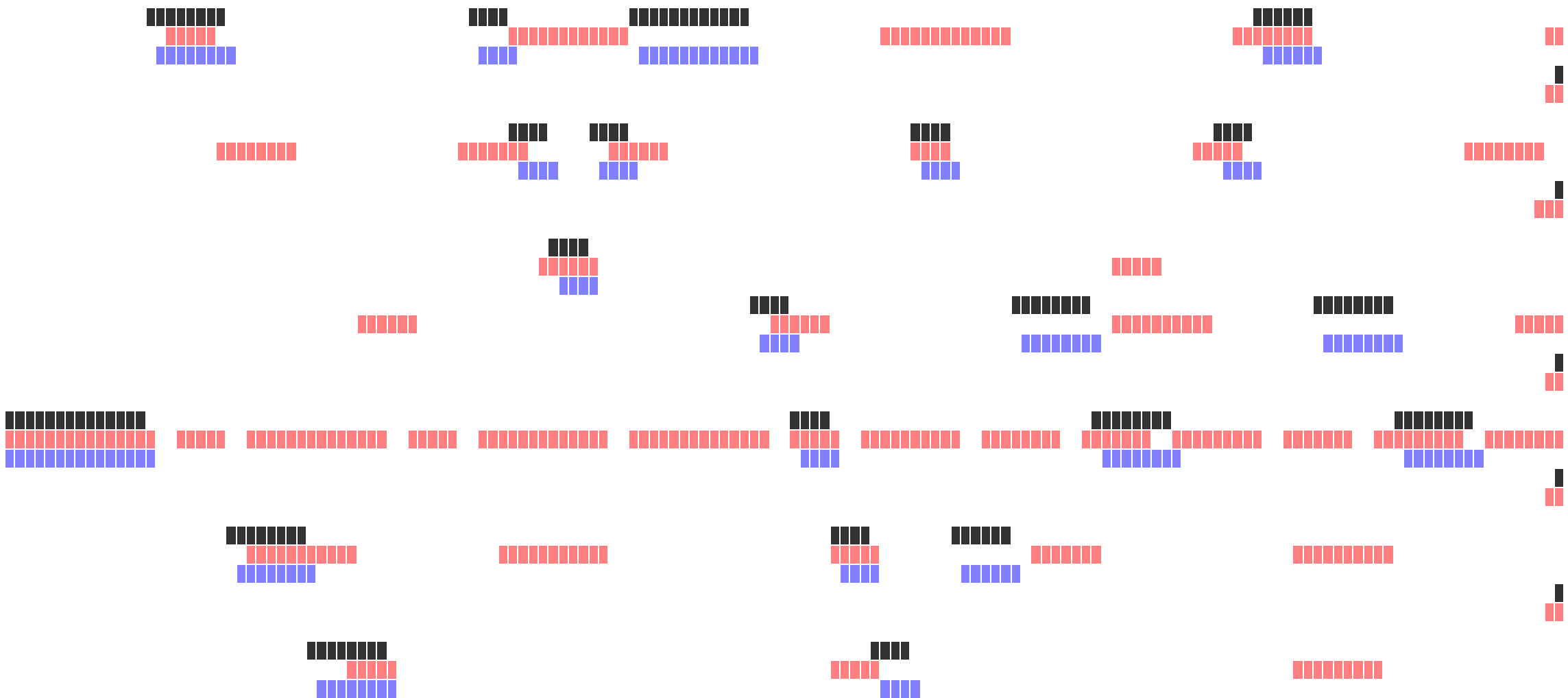}}
        \caption*{Music}
    \end{subfigure}\hspace{0.01\linewidth}
    \begin{subfigure}{0.32\linewidth}
        \fbox{\includegraphics[width=\textwidth]{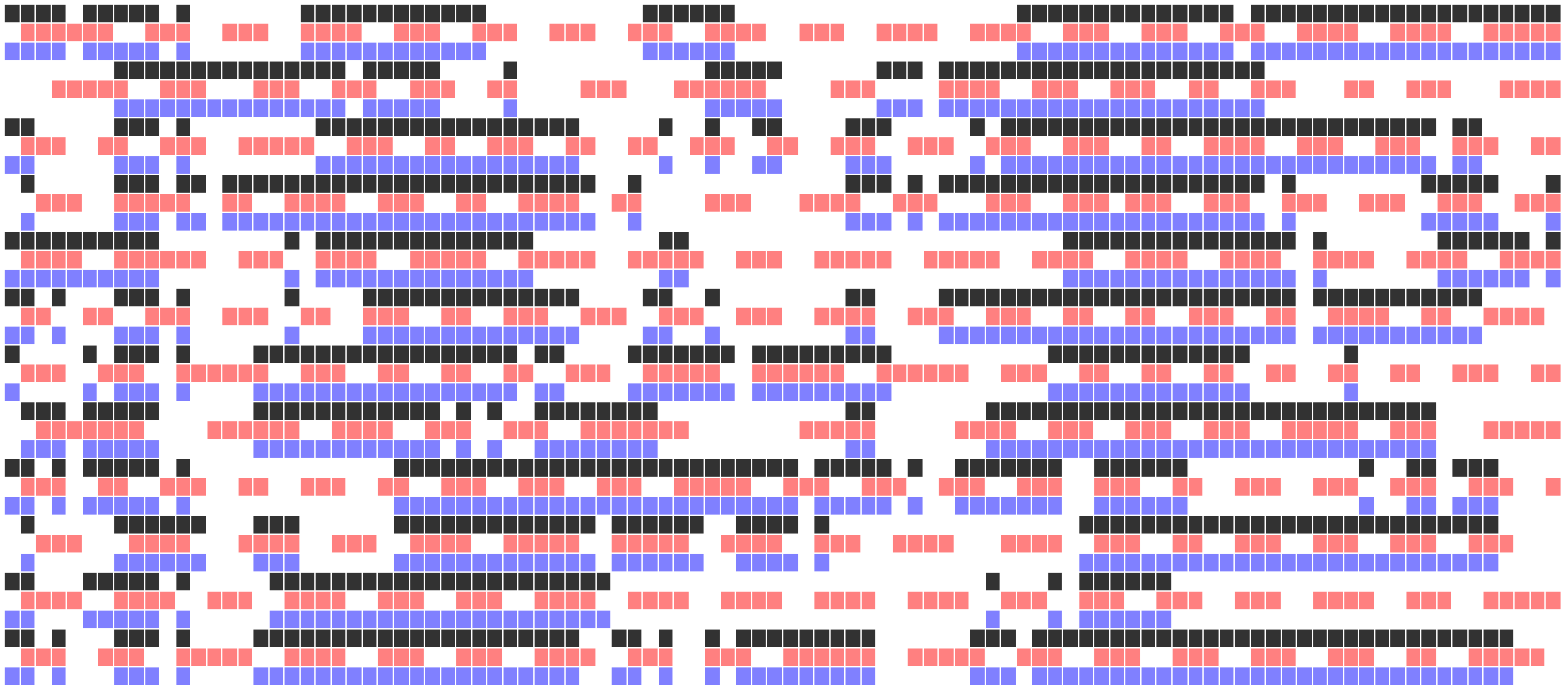}}
        \caption*{Stock}
    \end{subfigure}
    \caption{LTDDMs and LSTMs that have learned the three data sets after 200 epochs of training. Heartbeat shows the first 100 time steps on all 10 heartbeat streams. Music depicts the entirety of all 12 streams. Stock predictions correspond to the first 100 time steps of the first 10 streams.}\label{fig:samples}
\end{figure}

We first look at the quality of the models in terms of correct predictions at each time step. Tables~\ref{tab:results}-\ref{tab:results2} show the precision, recall, $F_1$ and accuracy of each model on each dataset. While LSTM leads in terms of precision, its recall performance goes from amazing (on Stock) to hilarious (on Heartbeat). A further analysis revealed that on Heartbeat, since the events are very sparse, LSTM simply chose not to respond most of the time, avoiding the 2 time steps cost of getting the timing wrong. As shown in Table~\ref{tab:results2}, LSTM does the same thing as the majority vote on this problem, and not much better than majority vote on Music. It is only on stocks, where the signal is ON about 60\% of the time, that LSTM does much better. This can also be observed on Fig.~\ref{fig:samples}. There is no clear winner between TDDM and LTDDM under these measures.

\vspace{-0.5cm}
\begin{table}
\caption{Performances of TDDM, LTDDM and LSTM on all three datasets. LSTM has better precision, but does not perform as well on recall except on Stock.}
\setlength{\tabcolsep}{16pt}
\renewcommand{\arraystretch}{1.3}
\begin{center}
    \resizebox{\textwidth}{!}{\begin{tabular}{c|ccc|ccc}
        \hline
         & \multicolumn{3}{c|}{Precision} & \multicolumn{3}{c}{Recall}  \\
         Dataset & TDDM & LTDDM & LSTM & TDDM& LTDDM & LSTM\\
         \hline
         Heartbeat & 0.23 & 0.19 & \textbf{0.52} & \textbf{0.36} & 0.31 & 0.03  \\
         Music & 0.32 & 0.34 & \textbf{0.48} & 0.68 & \textbf{0.75} & 0.48  \\
         Stock & 0.60 & 0.60 & \textbf{1.00} & 0.61 & 0.62 & \textbf{1.00} \\
         \hline
    \end{tabular}}
\end{center}
\label{tab:results}
\end{table}
\vspace{-1.5cm}

\begin{table}
\caption{Performances of TDDM, LTDDM and LSTM on all three datasets. Note how LSTM is identical or  to majority vote (always ON or always OFF) in Heartbeat and Music respectivly.}
\setlength{\tabcolsep}{16pt}
\renewcommand{\arraystretch}{1.3}
\begin{center}
    \resizebox{\textwidth}{!}{\begin{tabular}{c|ccc|cccc}
        \hline
         & \multicolumn{3}{c|}{$F_1$}  & \multicolumn{4}{c}{Accuracy}\\
         Dataset & TDDM & LTDDM & LSTM & TDDDM & LTDDM & LSTM & Majority Vote\\
         \hline
         Heartbeat & \textbf{0.28} & 0.24 & 0.06 & 0.80 & 0.79 & \textbf{0.89} & \textbf{0.89}\\
         Music & 0.43 & 0.46 & \textbf{0.48} & 0.82 & 0.84 & \textbf{0.98} & 0.93\\
         Stock & 0.60 & 0.61 & \textbf{1.00} & 0.53 & 0.54 & \textbf{1.00} & 0.60\\
         \hline
    \end{tabular}}
\end{center}
\label{tab:results2}
\end{table}
\vspace{-0.5cm}

The STE error of each model under each dataset is provided in Table~{\ref{tab:temporal}}. LTDDM had the lowest STE for the sparsest test, the Heartbeat dataset. This result is not surprising, since LTDDM minimizes its STE. Due to timescale invariance, LTDDM's timing error gradients will be largest at the beginning of learning. By performing gradient descent, LTDDM rapidly reduces its STE to find a local minimum, often within the first few epochs. When considering the timing error of these two models, it can be said that LTDDM learns sparse time series orders of magnitude faster than LSTM.

However, LTDDM did not have the lowest STE among models trained on the Music dataset. The learning model with the lowest STE for the Music dataset was LSTM. This was surprising, since LTDDM was designed specifically to learn these sorts of sequences and LSTM demonstrated significant difficulty learning to predict any of the events in the Heartbeat dataset. Further analysis revealed that the LSTM network is actually only repeating the input as output in Music, therefore, not making any predictions at all. The errors are almost all at the first time step the signal is ON and the first time step the signal is OFF. We have included the STE of such \textit{reactive} model in Table~\ref{tab:temporal}. This can be seen in Figure~\ref{fig:samples} and larger in Figure~\ref{fig:Music} in the appendix. Although LSTM was able to achieve the lowest temporal error for this test, it was not able to learn the timing of any of the events in the sequence. This result brings into question whether an LSTM model could ever learn the timing of notes in music for which these easily discovered and low error solutions exist.

\begin{figure}[t]
    \vspace{-1.5cm}
    \setlength{\fboxsep}{0pt}
    \setlength{\fboxrule}{1pt}
    \begin{subfigure}{\linewidth}
        \hspace{0.25\linewidth}\fcolorbox{black}{newGray}{\rule{0pt}{4pt}\rule{4pt}{0pt}}\quad TDDM
        \hspace{0.05\linewidth}\fcolorbox{black}{newRed}{\rule{0pt}{4pt}\rule{4pt}{0pt}}\quad LTDDM
        \hspace{0.05\linewidth}\fcolorbox{black}{newBlue}{\rule{0pt}{4pt}\rule{4pt}{0pt}}\quad LSTM
    \end{subfigure}\par\medskip
    \begin{subfigure}{0.32\linewidth}
        \includegraphics[width=\textwidth]{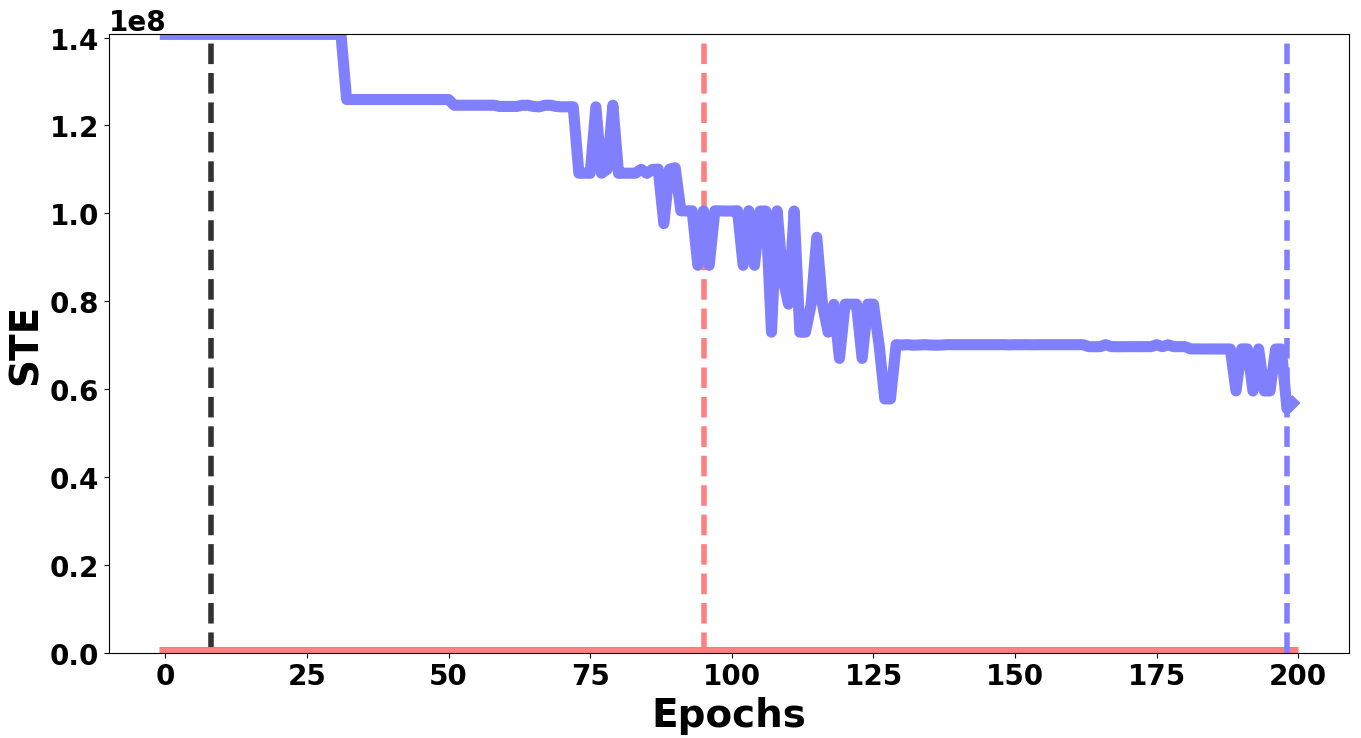}
        \caption*{Heartbeat}
    \end{subfigure}\hspace{0.01\linewidth}
    \begin{subfigure}{0.32\linewidth}
        \includegraphics[width=\textwidth]{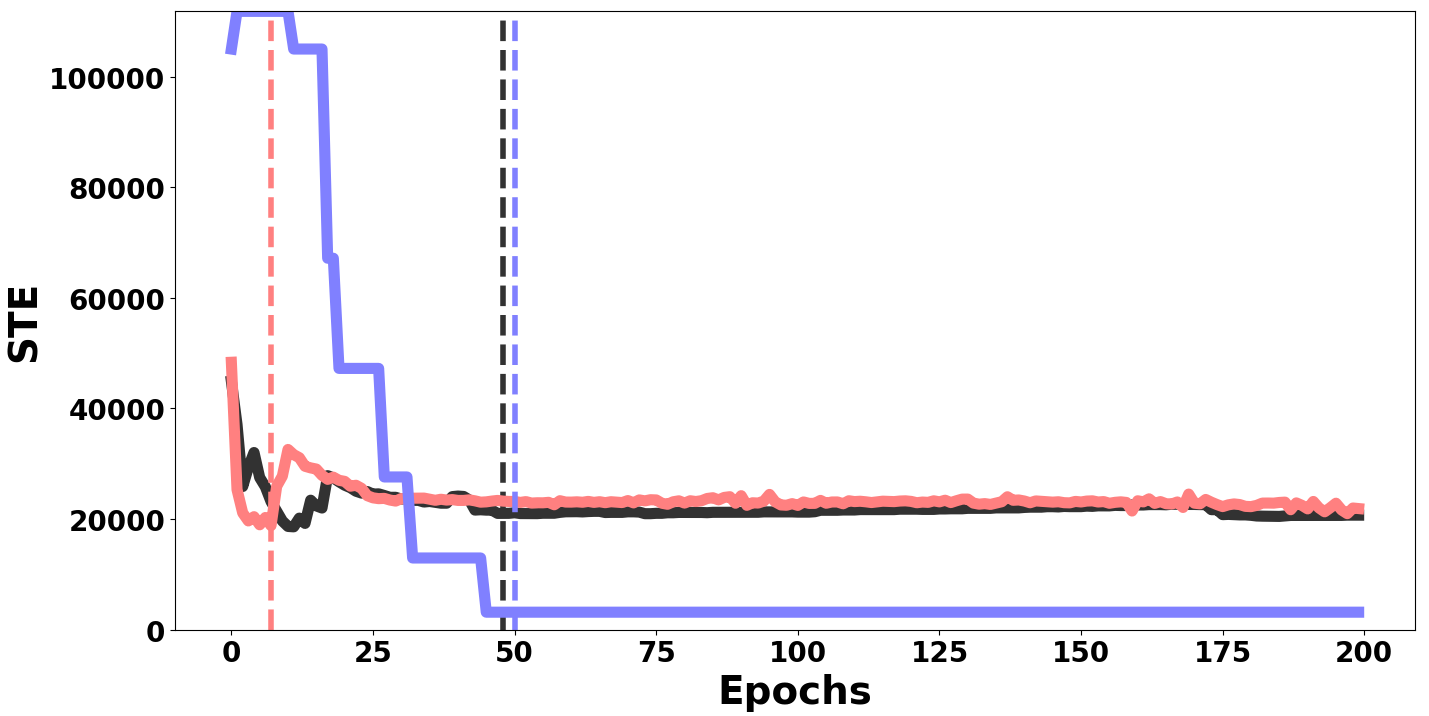}
        \caption*{Music}
    \end{subfigure}\hspace{0.01\linewidth}
    \begin{subfigure}{0.32\linewidth}
        \includegraphics[width=\textwidth]{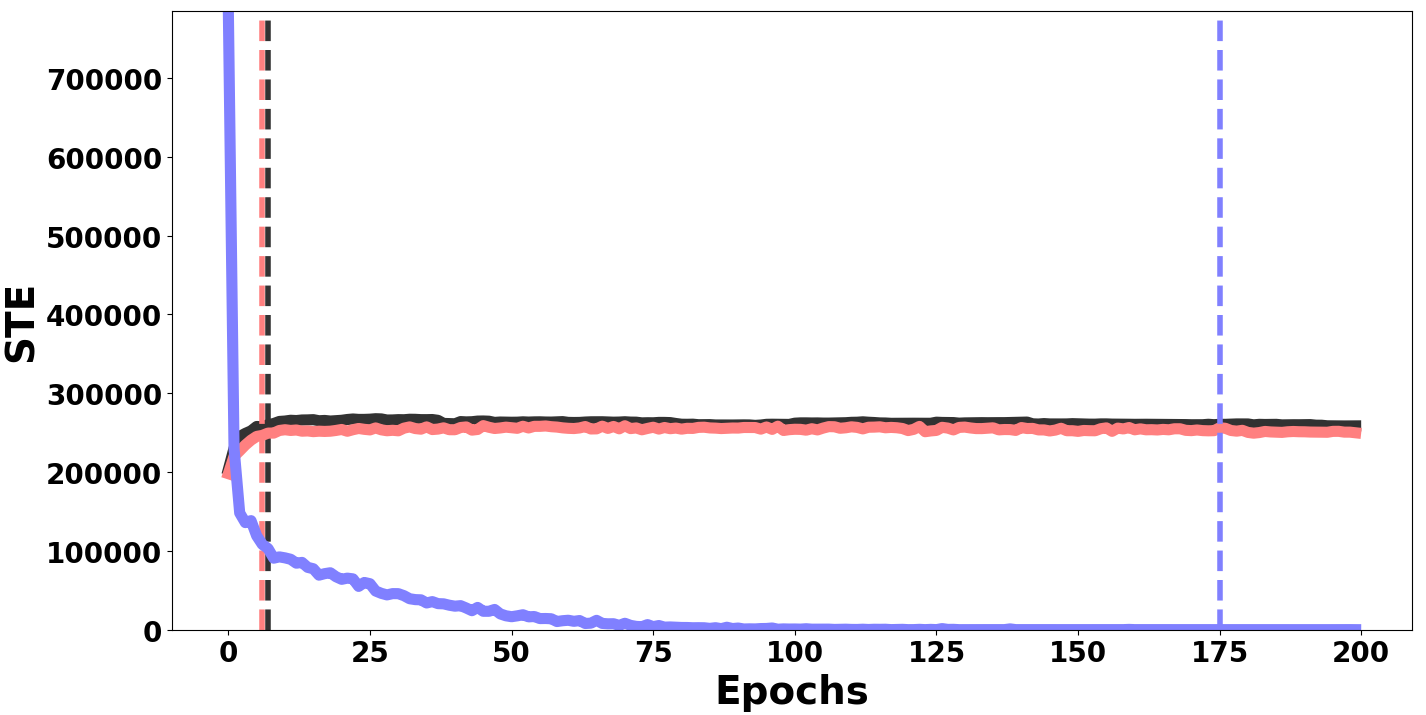}
        \caption*{Stock}
    \end{subfigure}
    \caption{The mean STE of all models on all problems in each data set over 200 epochs. Dashed vertical lines indicate the epoch at which each model's learning converged to a solution.}
    \label{fig:curves}
\end{figure}

In short, LSTM only truly learns something (and does it better than TDDM and LTDDM) in the cases where events (on ON state) are covering an important fraction of the steps. When events are rare or ON only for short periods of time, LSTM is quiet or simply replicates its input without any \textit{predictions}. Only TDDM and LTDDM are attempting to make predictions in those case. 
Finally, all models where compared for number of epochs to convergence (see Table~\ref{tab:temporal} and Fig.~\ref{fig:curves}). As expected, TDDM and LTDDM were always faster than LSTM.

\vspace{-.5cm}
\begin{table}[H]
    \setlength{\tabcolsep}{16pt}
    \renewcommand{\arraystretch}{1.3}
    \centering
    \resizebox{\textwidth}{!}{\begin{tabular}{c|cccc|ccc}
        \hline
        & \multicolumn{4}{c|}{STE}  & \multicolumn{3}{c}{Number of epochs to converge} \\ Dataset
        & TDDM & LTDDM & LSTM & Reactive & TDDM & LTDDM & LSTM\\
        \hline
        Heartbeat & 10,497.5 & 9,720.5 & 2,903,001.5 & \textbf{1,654.0} & \textbf{8} & 95 & 949 \\
        Music & 21,032.5 & 18,817.0 & \textbf{3,161.5} & \textbf{3,161.5} & 48 & \textbf{7} & 50\\
        Stock & 260,163.5 & 246,553.0 & \textbf{4.0} & 6,183.0 & 7 & \textbf{6} & 175\\
        \hline
    \end{tabular}}
    \caption{Temporal error (STE) of each algorithm after 200 epochs of training on each dataset. Note that on Music, LSTM behaves like the reactive model, which makes no predictions. TDDM and LTDDM learn much faster than LSTM.}
    \label{tab:temporal}
\end{table}

\section{Discussion}
In learning the timing and duration of events in a target sequence that exhibits rhythmic structure, LTDDM can learn much faster than LSTM by reducing its error in time, rather than its level of activation at each time step. Although not all time series data are of this form, or have the desired properties, some problems can be reduced to a two choice timing problem. This was demonstrated in the stock market data used in our experiment, wherein the change in stock price at each time step was restated as simply up or down. LTDDM's relatively poor performance on this data set indicates that it cannot find a low order function of composite period that describes the variability of stock prices through time. This change in perspective, from learning amplitude behaviour through time to learning timing behaviour, can lead to new insights into the temporal structure of time series data.

LTDDM outperformed LSTM in both its statistical, amplitude accuracy and its timing accuracy for the Heartbeat test. In the case of the heartbeat data set, LTDDM's timing accuracy was far superior to that of LSTM. However, despite its relatively close approximation to the timing of the target events, the statistical measures of LTDDM's performance are abysmal when compared with those of LSTM. This disconnect between LTDDM's apparently high timing accuracy and its low amplitude accuracy highlights the relative importance of these two measures in assessing the performance of models that learn timing. Regardless of the difference in time between the occurrence of a target event and the timing of a model's prediction, for two choice timing problems the model's amplitude error will be the same; at each time step it either got it right or wrong. For time series data with sparse target events, reducing amplitude error at each time step can strongly bias a predictor towards not predicting any events at all, as can be seen in LSTM's inability to predict the heartbeat data and many of the sequences in the music data. This approach to evaluating the performance of a predictor that has learned a timing task is insufficient, in that it is unable to express error in the model's timing. For problems where a predictor's behaviour can be thought of as pressing buttons, like pressing the keys of a piano or the buttons of a digital game controller, the predictor's timing error is most salient and better describes its performance.

\section{Conclusion}
We have summarized the mechanics and dynamics of what could become the basis for deep learning model for animal learning of timing and defined an extended model that enables learning of intermediate representations of conditioned stimuli. In learning both the timing and duration of target events, LTDDM can learn orders of magnitude faster than state of the art recurrent neural networks, like LSTM, by rapidly reducing its timing error. This shift in perspective, from reducing amplitude error at all times to reducing timing error at the occurrence of target events, enables LTDDM to accurately learn the timing and duration of events in sparse time series data with rhythmic structure.

\bibliographystyle{splncs04}
\bibliography{bib}

\begin{thebibliography}{10}
\providecommand{\url}[1]{\texttt{#1}}
\providecommand{\urlprefix}{URL }
\providecommand{\doi}[1]{https://doi.org/#1}

\bibitem{balsam_timing_2002}
Balsam, P.D., Drew, M.R., Yang, C.: Timing at the {Start} of {Associative}
  {Learning}. Learning and Motivation  \textbf{33}(1),  141--155 (Feb 2002).
  \doi{10.1006/lmot.2001.1104}

\bibitem{BlumRivest1992}
Blum, A.L., Rivest, R.L.: Training a 3-node neutral network is {NP}-complete.
  Neural Networks  \textbf{5},  117--127 (1992)

\bibitem{cai_dtwnet_2019}
Cai, X., Xu, T., Yi, J., Huang, J., Rajasekaran, S.: {DTWNet}: a {Dynamic}
  {Time} {Warping} {Network}. In: Wallach, H., Larochelle, H., Beygelzimer, A.,
  Alché-Buc, F.d., Fox, E., Garnett, R. (eds.) Advances in {Neural}
  {Information} {Processing} {Systems} 32, pp. 11640--11650. Curran Associates,
  Inc. (2019)

\bibitem{deverett_interval_2019}
Deverett, B., Faulkner, R., Fortunato, M., Wayne, G., Leibo, J.Z.: Interval
  timing in deep reinforcement learning agents. In: Wallach, H., Larochelle,
  H., Beygelzimer, A., Alché-Buc, F.d., Fox, E., Garnett, R. (eds.) Advances
  in {Neural} {Information} {Processing} {Systems} 32, pp. 6686--6695. Curran
  Associates, Inc. (2019)

\bibitem{gallistel_time_2000}
Gallistel, C.R., Gibbon, J.: Time, rate, and conditioning. Psychological Review
   \textbf{107}(2), ~289 (May 2000). \doi{10.1037/0033-295X.107.2.289},
  publisher: US: American Psychological Association

\bibitem{gers_learning_2003}
Gers, F.A., Schraudolph, N.N., Schmidhuber, J.: Learning precise timing with
  lstm recurrent networks. The Journal of Machine Learning Research
  \textbf{3},  115--143 (Mar 2003). \doi{10.1162/153244303768966139}

\bibitem{gibbon_scalar_1977}
Gibbon, J.: Scalar expectancy theory and {Weber}'s law in animal timing.
  Psychological Review  \textbf{84}(3),  279--325 (1977).
  \doi{10.1037/0033-295X.84.3.279}

\bibitem{hawthorne_enabling_2019}
Hawthorne, C., Stasyuk, A., Roberts, A., Simon, I., Huang, C.Z.A., Dieleman,
  S., Elsen, E., Engel, J., Eck, D.: Enabling {Factorized} {Piano} {Music}
  {Modeling} and {Generation} with the {MAESTRO} {Dataset}. arXiv:1810.12247
  [cs, eess, stat]  (Jan 2019), \url{http://arxiv.org/abs/1810.12247}, arXiv:
  1810.12247

\bibitem{hochreiter_long_1997}
Hochreiter, S., Schmidhuber, J.: Long {Short}-{Term} {Memory}. Neural
  Computation  \textbf{9}(8),  1735--1780 (Nov 1997).
  \doi{10.1162/neco.1997.9.8.1735}

\bibitem{huang_music_2018}
Huang, C.Z.A., Vaswani, A., Uszkoreit, J., Shazeer, N., Simon, I., Hawthorne,
  C., Dai, A.M., Hoffman, M.D., Dinculescu, M., Eck, D.: Music {Transformer}.
  arXiv:1809.04281 [cs, eess, stat]  (Dec 2018),
  \url{http://arxiv.org/abs/1809.04281}, arXiv: 1809.04281

\bibitem{le_guen_shape_2019}
Le~Guen, V., Thome, N.: Shape and {Time} {Distortion} {Loss} for {Training}
  {Deep} {Time} {Series} {Forecasting} {Models}. In: Wallach, H., Larochelle,
  H., Beygelzimer, A., Alché-Buc, F.d., Fox, E., Garnett, R. (eds.) Advances
  in {Neural} {Information} {Processing} {Systems} 32, pp. 4189--4201. Curran
  Associates, Inc. (2019)

\bibitem{luzardo_2013}
Luzardo, A., Ludvig, E.A., Rivest, F.: An adaptive drift-diffusion model of
  interval timing dynamics. Behavioural Processes  \textbf{95},  90--99 (May
  2013). \doi{10.1016/j.beproc.2013.02.003}

\bibitem{luzardo_driftdiffusion_2017}
Luzardo, A., Rivest, F., Alonso, E., Ludvig, E.A.: A drift–diffusion model of
  interval timing in the peak procedure. Journal of Mathematical Psychology
  \textbf{77},  111--123 (Apr 2017). \doi{10.1016/j.jmp.2016.10.002}

\bibitem{rivest_adaptive_2011}
Rivest, F., Bengio, Y.: Adaptive {Drift}-{Diffusion} {Process} to {Learn}
  {Time} {Intervals}. arXiv:1103.2382 [q-bio]  (Mar 2011), arXiv: 1103.2382

\bibitem{rivest_new_2020}
Rivest, F., Kohar, R.: A {New} {Timing} {Error} {Cost} {Function} for {Binary}
  {Time} {Series} {Prediction}. IEEE Transactions on Neural Networks and
  Learning Systems  \textbf{31}(1),  174--185 (Jan 2020).
  \doi{10.1109/TNNLS.2019.2900046}

\bibitem{Simenetal2011}
Simen, P., Balci, F., de~Souza, L., Cohen, J.D., Holmes, P.: A model of
  interval timing by neural integration. Journal of Neuroscience
  \textbf{31}(25),  9238--9253 (2011). \doi{10.1523/JNEUROSCI.3121-10.2011}

\bibitem{tavanaei_deep_2019}
Tavanaei, A., Ghodrati, M., Kheradpisheh, S.R., Masquelier, T., Maida, A.: Deep
  learning in spiking neural networks. Neural Networks  \textbf{111},  47--63
  (Mar 2019). \doi{10.1016/j.neunet.2018.12.002}

\bibitem{trappenberg_fundamentals_2010}
Trappenberg, T.P.: Fundamentals of computational neuroscience. Oxford
  University Press, Oxford; New York (2010), oCLC: 804910180

\end{thebibliography}
\vfill

\appendix
\section*{Appendix}
\begin{figure}[H]
    \begin{subfigure}{\linewidth}
        \hspace{0.2\linewidth}\fcolorbox{black}{newGray}{\rule{0pt}{4pt}\rule{4pt}{0pt}}\quad Ground Truth
        \hspace{0.05\linewidth}\fcolorbox{black}{newRed}{\rule{0pt}{4pt}\rule{4pt}{0pt}}\quad LTDDM
        \hspace{0.05\linewidth}\fcolorbox{black}{newBlue}{\rule{0pt}{4pt}\rule{4pt}{0pt}}\quad LSTM
    \end{subfigure}\par\medskip
    \begin{subfigure}{\linewidth}
        \includegraphics[width=\textwidth,height=0.25\textheight]{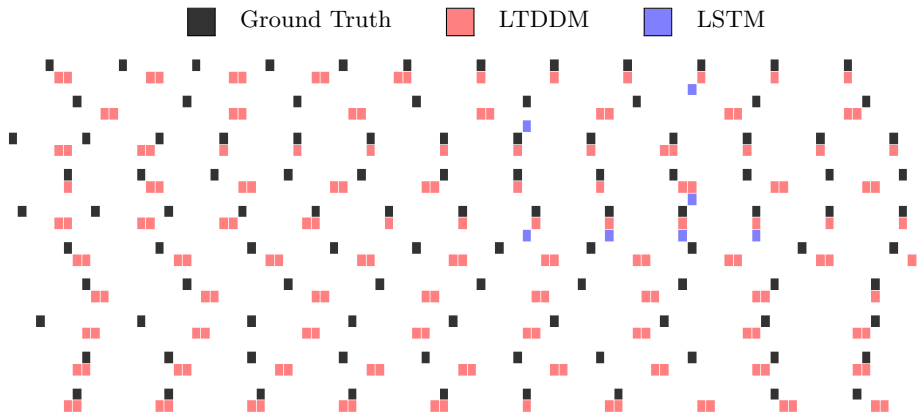}
    \end{subfigure}
    \caption{A LTDDM and a LSTM that have learned to predict a human heartbeat. The figure shows the first 100 time steps of all sequences. Due to the sparsity in target events, LSTM prefers to make fewer predictions to minimize its binary cross entropy.}
\end{figure}
\begin{figure}
    \begin{subfigure}{\linewidth}
        \hspace{0.2\linewidth}\fcolorbox{black}{newGray}{\rule{0pt}{4pt}\rule{4pt}{0pt}}\quad Ground Truth
        \hspace{0.05\linewidth}\fcolorbox{black}{newRed}{\rule{0pt}{4pt}\rule{4pt}{0pt}}\quad LTDDM
        \hspace{0.05\linewidth}\fcolorbox{black}{newBlue}{\rule{0pt}{4pt}\rule{4pt}{0pt}}\quad LSTM
    \end{subfigure}\par\medskip
    \begin{subfigure}{\linewidth}
        \includegraphics[width=\textwidth,height=0.25\textheight]{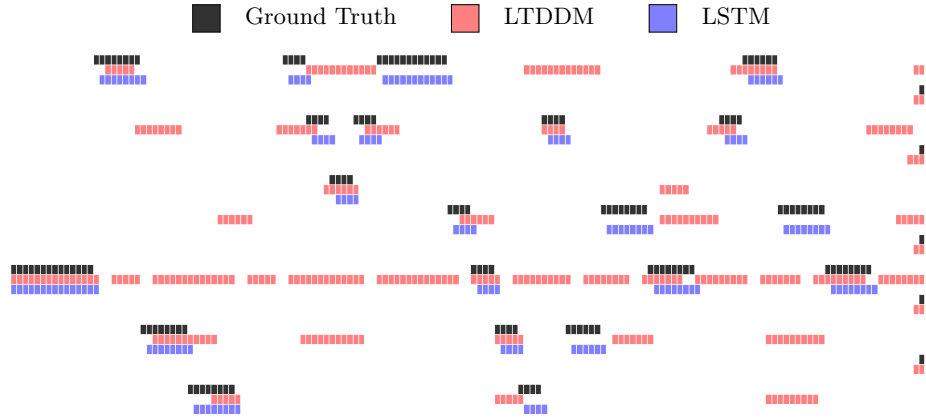}
    \end{subfigure}
    \caption{A LTDDM and a LSTM that have learned to play \textit{Die Naehterin's sass a Naehterin und sie naeht}. LTDDM makes many predictions with some timing error. LSTM makes predictions with perfect precision, in only those streams where notes occur before the end of the sequence. For the other sequences it fails to make any predictions at all.}
    \label{fig:Music}
\end{figure}
\begin{figure}
    \begin{subfigure}{\linewidth}
        \hspace{0.2\linewidth}\fcolorbox{black}{newGray}{\rule{0pt}{4pt}\rule{4pt}{0pt}}\quad Ground Truth
        \hspace{0.05\linewidth}\fcolorbox{black}{newRed}{\rule{0pt}{4pt}\rule{4pt}{0pt}}\quad LTDDM
        \hspace{0.05\linewidth}\fcolorbox{black}{newBlue}{\rule{0pt}{4pt}\rule{4pt}{0pt}}\quad LSTM
    \end{subfigure}\par\medskip
    \begin{subfigure}{\linewidth}
        \includegraphics[width=\textwidth,height=0.25\textheight]{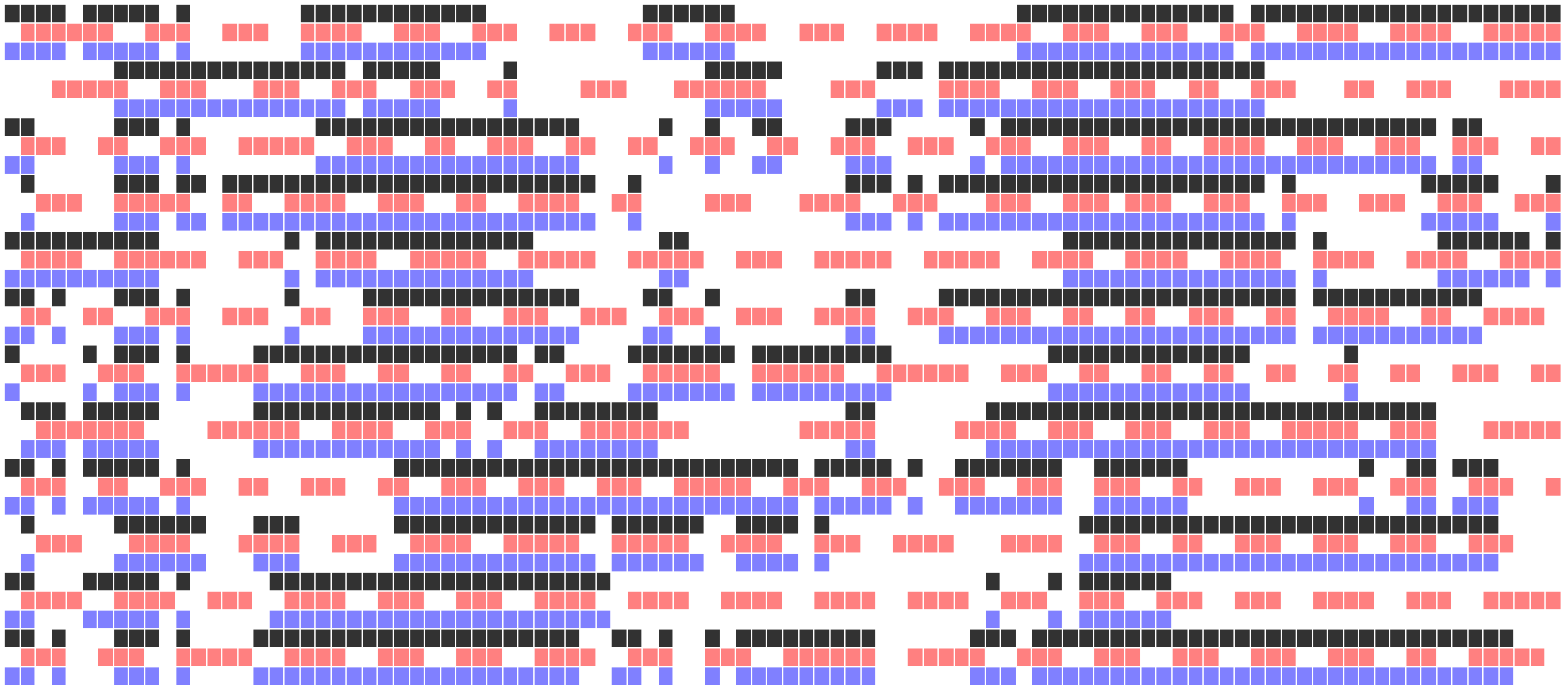}
    \end{subfigure}
    \caption{A LTDDM and a LSTM that have learned to predict whether the price of a stock went up or down. The figure portrays the first 100 time steps of the first 10 sequnces. LTDDM has difficulty learning sequences with little rhythmic structure.}
\end{figure}

\begin{figure}
    \begin{subfigure}{\linewidth}
        \hspace{0.25\linewidth}\fcolorbox{black}{newGray}{\rule{0pt}{4pt}\rule{4pt}{0pt}}\quad TDDM
        \hspace{0.05\linewidth}\fcolorbox{black}{newRed}{\rule{0pt}{4pt}\rule{4pt}{0pt}}\quad LTDDM
        \hspace{0.05\linewidth}\fcolorbox{black}{newBlue}{\rule{0pt}{4pt}\rule{4pt}{0pt}}\quad LSTM
    \end{subfigure}
    \begin{subfigure}{\linewidth}
        \includegraphics[width=\textwidth]{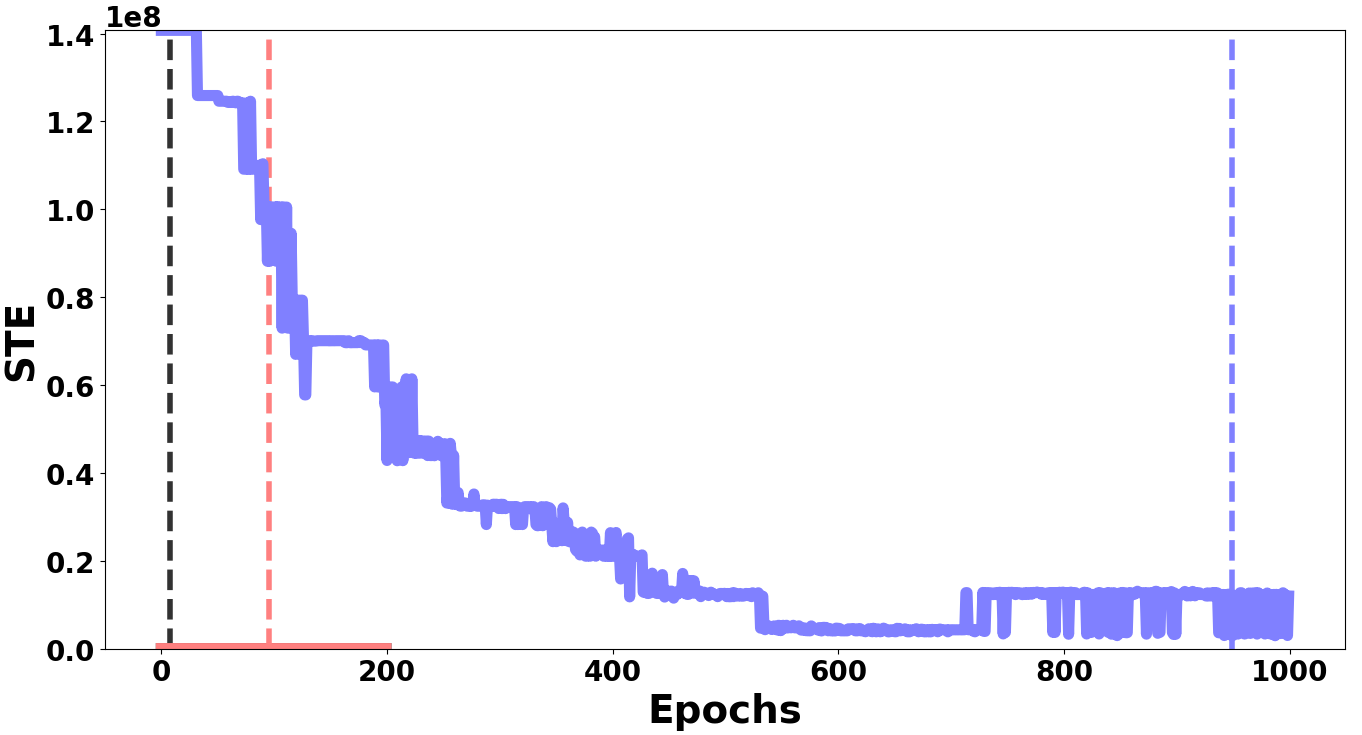}
    \end{subfigure}
    \caption{The mean STE of all models on all streams in the Heartbeat data set over 200 epochs for TDDM and LTDDM and 1000 epochs for LSTM.}
\end{figure}

\begin{figure}
    \begin{subfigure}{\linewidth}
        \hspace{0.25\linewidth}\fcolorbox{black}{newGray}{\rule{0pt}{4pt}\rule{4pt}{0pt}}\quad TDDM
        \hspace{0.05\linewidth}\fcolorbox{black}{newRed}{\rule{0pt}{4pt}\rule{4pt}{0pt}}\quad LTDDM
        \hspace{0.05\linewidth}\fcolorbox{black}{newBlue}{\rule{0pt}{4pt}\rule{4pt}{0pt}}\quad LSTM
    \end{subfigure}
    \begin{subfigure}{\linewidth}
        \includegraphics[width=\textwidth]{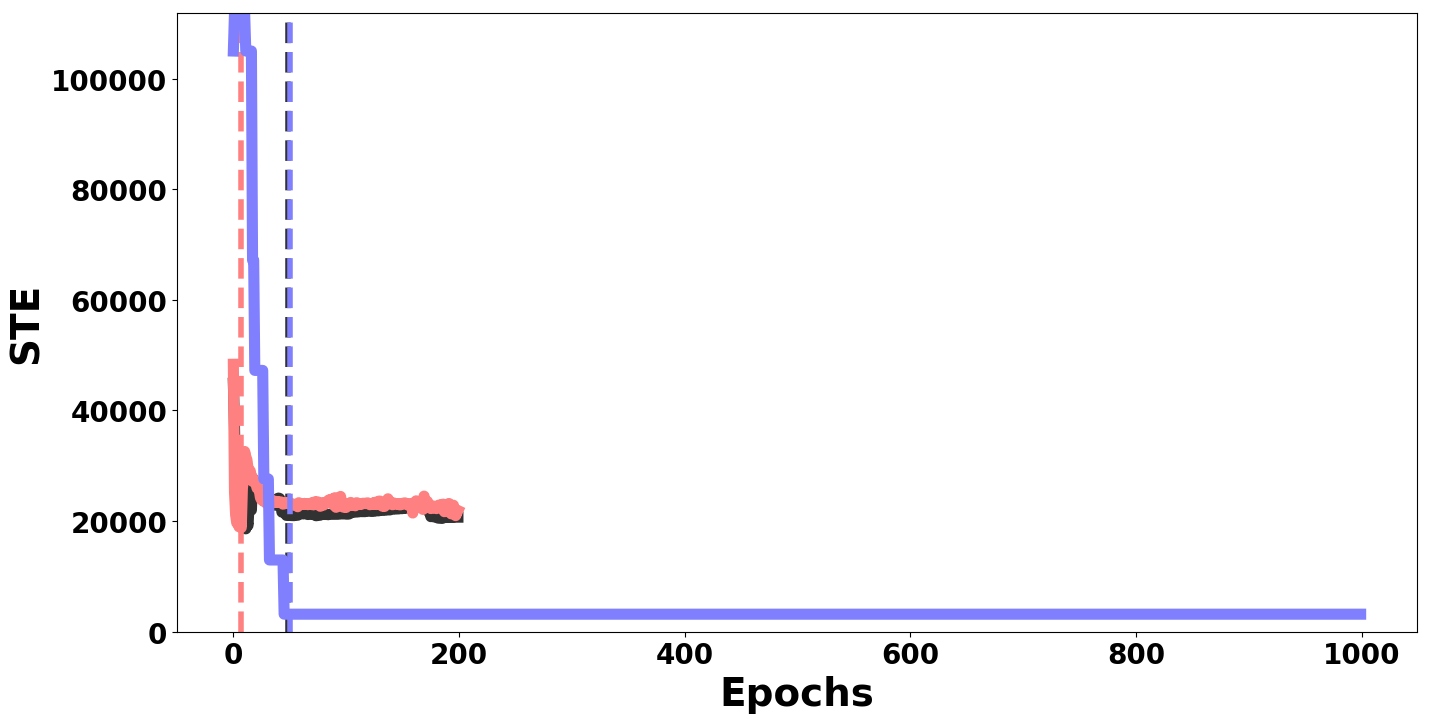}
    \end{subfigure}
    \caption{The mean STE of all models on all streams in the Music data set over 200 epochs for TDDM and LTDDM and 1000 epochs for LSTM.}
\end{figure}

\begin{figure}
    \begin{subfigure}{\linewidth}
        \hspace{0.25\linewidth}\fcolorbox{black}{newGray}{\rule{0pt}{4pt}\rule{4pt}{0pt}}\quad TDDM
        \hspace{0.05\linewidth}\fcolorbox{black}{newRed}{\rule{0pt}{4pt}\rule{4pt}{0pt}}\quad LTDDM
        \hspace{0.05\linewidth}\fcolorbox{black}{newBlue}{\rule{0pt}{4pt}\rule{4pt}{0pt}}\quad LSTM
    \end{subfigure}
    \begin{subfigure}{\linewidth}
        \includegraphics[width=\textwidth]{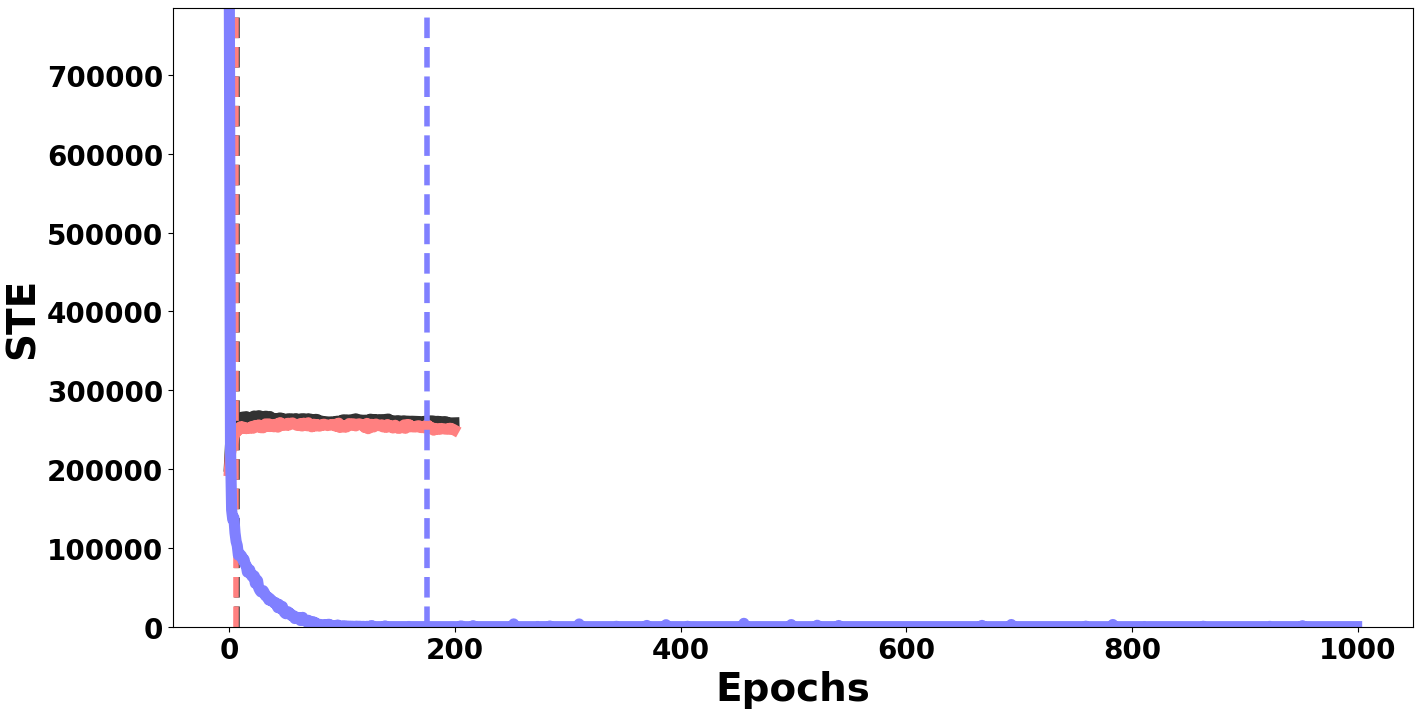}
    \end{subfigure}
    \caption{The mean STE of all models on all streams in the Stock data set over 200 epochs for TDDM and LTDDM and 1000 epochs for LSTM.}
\end{figure}

\end{document}